\documentclass[journal=gmj]{CUP-JNL-DTM}%

\usepackage{makecell}
\usepackage{graphicx}
\usepackage{multicol,multirow}
\usepackage{amsmath,amssymb,amsfonts}
\usepackage{mathrsfs}
\usepackage{amsthm}
\usepackage{rotating}
\usepackage{siunitx}
\usepackage{appendix}
\usepackage{ifpdf}
\usepackage{hyperref}
\usepackage[T1]{fontenc}
\usepackage{newtxtext}
\usepackage{newtxmath}
\usepackage{graphicx}
\usepackage{textcomp}
\usepackage{xcolor}
\usepackage{lipsum}
\usepackage{comment}
\usepackage{xspace}

\theoremstyle{definition}

\numberwithin{equation}{section}

\newcommand{\metric}{\textsc{Space}\xspace}
\newcommand{\methodBench}{\textsc{SynopticBench}\xspace}
\newcommand{\ie}{\textit{i.e., \xspace}}
\newcommand{\eg}{\textit{e.g., \xspace}}
\newcommand{\xhdr}[1]{\vspace{0em}\noindent{{\bf #1.}}}
\newcommand{\std}[1]{$\mathbin{\scriptstyle\pm#1}$}

\jname{Data/Math}
\articletype{Data Paper}
\jyear{2026}

\begin{document}

\begin{Frontmatter}

\title[Article Title]{\methodBench: Evaluating Vision-Language Models on Generating Weather Forecast Discussions of the Future}

\author[1,2]{Timothy B. Higgins}
\author[1,2]{Antonios Mamalakis}
\author[2]{Chirag Agarwal}

\authormark{Higgins \textit{et al}.}

\address[1]{\orgdiv{Department of Environmental Sciences}, \orgname{University of Virginia}, \orgaddress{\city{Charlottesville}, \postcode{22903}, \state{Virginia},  \country{United States}}}

\address[2]{\orgdiv{School of Data Science}, \orgname{University of Virginia}, \orgaddress{\city{Charlottesville}, \postcode{22903}, \state{Virginia},  \country{United States}}}

\authormark{Higgins et al.}

\keywords{Multimodality, Numerical Weather Prediction, Vision Language Model, Weather Forecasting}

\keywords[MSC Codes]{\codes[Primary]{CODE1}; \codes[Secondary]{CODE2, CODE3}}

\abstract{
Recent advances in visual-language models (VLMs) have led to significant improvements in a plethora of complex multimodal tasks like image captioning, report generation, and visual perception. However, generating text from meteorological data is highly challenging because the atmosphere is a chaotic system that is rapidly changing at various spatial and temporal scales. Given the complexity of atmospheric phenomena, it is critical to verifiably quantify the effectiveness of existing VLMs on weather forecasting data. In this work, we present \methodBench, a high-quality dataset consisting of 1,367,041 text samples of Area Forecast Discussions created by the National Weather Service over the continental United States paired to images of 500mb geopotential height, 2 meter temperature, and 850mb wind velocity in weather forecasts. We also present Synoptic Phenomena Alignment and Coverage Evaluation (\metric), a novel evaluation framework that can be used to effectively estimate the quality of text descriptions of synoptic weather phenomena. Extensive experiments on generating forecast discussions using state-of-the-art VLMs show the sensitivity of existing evaluation metrics in this domain and enable further exploration into synoptic weather and climate text generation.
}

\end{Frontmatter}

\section*{Impact Statement}
To the best of our knowledge, this is the first study to generate text discussions from images of numerical weather forecasts. We introduce i) the largest benchmark dataset, comprising 1,367,041 image-text samples, to run experiments that explore the potential to generate text discussions from numerical weather prediction; and ii) a methodology for evaluating generated text discussions of synoptic weather and climate.

\localtableofcontents

\section{Introduction}
Recent advances in vision-language models (VLMs) have enabled substantial progress on complex multimodal tasks such as image captioning, document understanding, visual question answering, and multimodal summarization. As these models improve in capability and are increasingly deployed or proposed for deployment in high-stakes settings, there is a growing need for evaluation frameworks that go beyond surface-level text similarity and instead quantify whether generated language is faithful, spatially grounded, and domain-relevant.

To this end, weather forecasting is a particularly consequential and technically challenging high-stakes domain. Numerical Weather Prediction (NWP) is deeply integrated into business operations, agriculture, government decision-making, public safety, and daily life for billions of people \cite{jones_potential_2000,jain_big_2017,uccellini_evolving_2019,ukhurebor_precision_2022,noauthor_valuing_2015}. Vast observation networks and reanalysis products have enabled continual improvements in model development, data assimilation, and assessments of predictive reliability \citep{hersbach_era5_2020,randles_merra-2_2017}. However, atmospheric dynamics are chaotic and span a wide range of spatial and temporal scales, which is why forecast verification relies on many complementary skill metrics (\eg Anomaly Correlation Coefficient, RMSE, MAE, and skill scores) to characterize different aspects of performance \citep{brady_climpred_2021}. In contrast, despite the operational importance of narrative forecasting products, methods to verifiably quantify the reliability of textual representations of atmospheric states and evolution \textbf{remain limited}.

Recent years have seen a growing body of work that explores multimodal learning for atmospheric data, including fine-tuning frontier VLMs and dataset creation for meteorological question answering (QA) and event understanding. Examples include CLLMate~\citep{li_cllmate_2024}, which pairs extreme-event news with reanalysis for QA; RadarQA~\citep{he_radarqa_2025}, which introduces radar annotations and heuristics for VLM fine-tuning; and hazard-focused datasets such as SEVIR~\citep{veillette_sevir_2020} and GridRad-Severe~\citep{murphy_development_2023}. Additional efforts pair structured meteorological inputs with textual outputs for QA-oriented benchmarks, such as MeteorPred~\citep{tang_meteorpred_2025}, Climate IQA~\citep{chen_climateiqa_2025}, Zephyrus~\citep{varambally_zephyrus_2025}, and WeatherQA~\citep{ma_weatherqa_2024}. While these studies demonstrate the promise of multimodal models for atmospheric applications, direct comparisons across benchmarks are often confounded by rigid constraints on target outputs, spatial scales, forecast horizons, and training data choices. Moreover, many evaluations emphasize question-answering accuracy or rely heavily on traditional text metrics (\textsc{Bleu}, \textsc{Rouge}, \textsc{Meteor}, \textsc{BertScore}) and/or LLM-judge scoring, which can be insufficient for assessing whether generated discussions correctly identify synoptic phenomena, capture their spatial extent, and reflect scale-dependent relevance to a forecast domain.

\xhdr{Present work} To address all the aforementioned gaps, we introduce a new benchmark and evaluation methodology designed specifically for the generation and verification of synoptic-scale forecast discussions. We present \methodBench, a large-scale data set comprising 1,367,041 National Weather Service Area Forecast Discussions paired with forecast images of 500mb geopotential height, 2m temperature and 850mb wind velocity over the continental United States. In contrast to existing benchmarks, \methodBench is the first benchmark to evaluate NWP meteorological samples across both mesoscale and synoptic spatial scales for all weather (see Table~\ref{tab:table1}). Critically, we also propose Synoptic Phenomena Alignment and Coverage Evaluation (\metric), an evaluation framework that explicitly accounts for the scale-dependent nature of atmospheric phenomena and the differing spatial precision required across variables (\eg precipitation versus pressure patterns). \metric is designed to evaluate distinct phenomena separately while incorporating spatial coverage and relevance, enabling more faithful estimation of a VLM's skill in generating meteorological text. Building on \methodBench, we train models for synoptic discussion generation and fine-tune widely used VLMs. Our results show that the ability of the models to conduct physical reasoning is improved from finetuning.
The key goal of \methodBench is to advance the study and understanding of vision-language models in weather forecasting. Therefore, we share the dataset, metric, and fine-tuned weights of a range of VLMs to support reproducible research and more rigorous, domain-appropriate evaluation of VLMs in weather forecasting.
\begin{table}
\centering
\small
\setlength{\tabcolsep}{8pt}
\renewcommand{\arraystretch}{1.1}
\begin{tabular}{lcccc}
\toprule
  Dataset  & \makecell{Meteorological \\Data} &  Text Size & Spatial Scale & Target\\
\midrule
   GridRad-Severe ~\cite{murphy_development_2023}  & RS & x & \makecell{SC, MC, MS} & Extreme Events \\
   SEVIR~\cite{veillette_sevir_2020}
  & RS  & x & MS & Extreme Events\\
   WeatherQA~\cite{ma_weatherqa_2024} & \makecell{Reanalysis} & 8,000 & MS & Extreme Events\\
   ClimateIQA~\cite{chen_climateiqa_2025} & Reanalysis & 762,120 & \makecell{MS, Synoptic} & Extreme Events\\
   CLLMate~\cite{li_cllmate_2024}  & Reanalysis & 41,000 & MS & Extreme Events\\
   RadarQA~\cite{he_radarqa_2025}  & RS &  69,000 & Mesoscale & All Weather\\
   MP-Bench~\cite{tang_meteorpred_2025}  & Reanalysis & 421,363 & MS & Extreme Events\\
   Zephyrus Bench~\cite{varambally_zephyrus_2025}  & Reanalysis & 2,158 & \makecell{MS, Synoptic, \\ Global} & All Weather\\
   \methodBench & \textbf{NWP}  &  \textbf{1,367,041} & \textbf{\makecell{MS and Synoptic}} & \textbf{All Weather}\\
 \bottomrule
 \end{tabular}
  \label{tab:table1}
  \vspace{0.05in}
\caption{Comparison between \methodBench and existing work using multimodal datasets for atmospheric data. Here, we present the type of meteorological data (RS$\to$Remote Sensing; NWP$\to$Numerical Weather Prediction), the number of text samples, the relevant spatial scales (SC$\to$Single-cell; MC$\to$Multi-cell; MS$\to$Mesoscale), and the target of each study. \methodBench is the first large-scale benchmark that generates text from NWP and contains 1,367,041 text samples
}
 \end{table}

\section{\methodBench}
Here, we detail the dataset building pipeline (Sec.~\ref{sec:pipeline}), data collection (Sec.~\ref{sec:collection}), data preprocessing (Sec.~\ref{sec:preprocessing}), and our proposed \metric metric (Sec.~\ref{sec:metric}).

\subsection{Dataset Building Pipeline}
\label{sec:pipeline}

One of the largest bottlenecks that stunts progress for multimodal models for atmospheric data is the scarcity of high-quality text that can be paired with numerical data. We propose using Area Forecast Discussions (AFDs) created by the National Weather Service (NWS) as a potential solution to the multimodal data problem. The AFD text data are publicly available and were extracted from the Iowa Environmental Mesonet\footnote{\href{mesonet.agron.iastate.edu/wx/afos/list.phtml}{https://mesonet.agron.iastate.edu/wx/afos/list.phtml}}.
AFDs exist at NWS stations across the United States and are typically issued multiple times per day. 
They reliably describe weather forecasts to an exceptionally high degree of detail, often informing the public about the weather a week in advance \cite{noauthor_completing_2006}.
In addition, they describe numerous physical variables, vertical levels, spatial scales, and processes. AFDs also typically include high-quality descriptions of the temporal evolution of weather phenomena, creating potential for training models to generate text that describes how atmospheric data changes over time.

\subsection{Dataset Collection}
\label{sec:collection}
\begin{figure}[h!]
 \centerline{\includegraphics[width=0.81\textwidth]{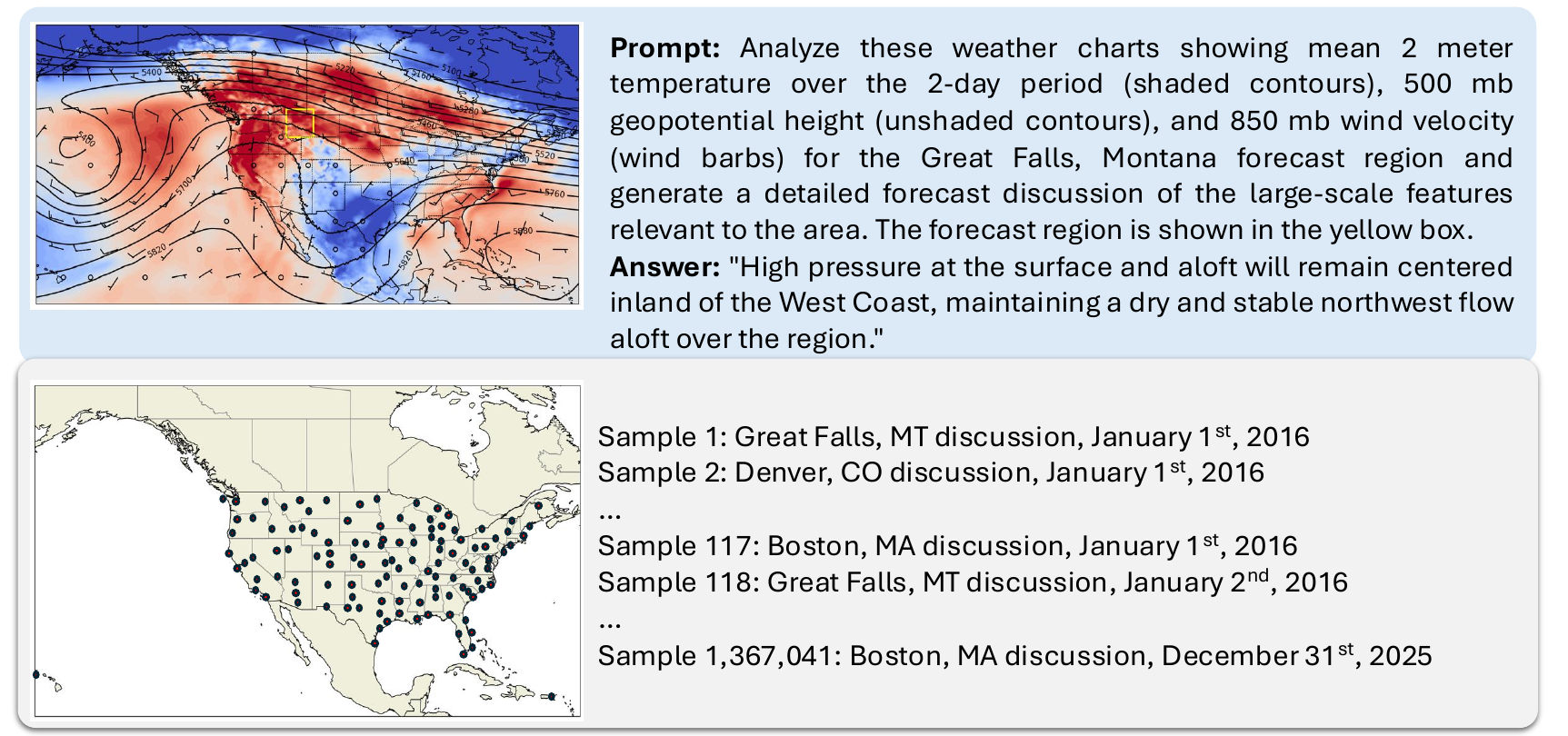}}
  \caption{An example case of a single sample from the training set (top panel). Each training sample image has a yellow box indicating the location of the discussion. The example answer is a filtered AFD. All of the locations used in the discussions are shown in the bottom panel. The format of the training samples is also shown, with 117 AFDs matched to each forecast}\label{Fig1}
\end{figure}

We create \methodBench by pairing the National Centers for Environmental Prediction (NCEP) Global Forecast System (GFS) model \citep{national_centers_for_environmental_predictionnational_weather_servicenoaaus_department_of_commerce_ncep_2015} to AFDs. This data set contains GFS forecasts issued four times per day from January 1, 2016$\to$December 31, 2025 (14,159 total forecasts) paired with \textbf{1,367,041} AFD text samples. Each GFS forecast consists of 2m temperature, precipitation rate, 500mb geopotential height, 850mb zonal wind, and 850 mb meridional wind is matched to the nearest AFD in time (always within several hours) for each location. We pair each forecast with 117 text samples, as there are 117 NWS station locations used in the study (Figure 1). This creates consistency among the text samples for descriptions of the primary synoptic features occurring over the US for a given forecast.

\subsection{Preprocessing}
\label{sec:preprocessing}

Before we begin training and evaluating vision-language models, we apply several data preprocessing steps to both AFDs and GFS forecasts. During preprocessing, we first discard all lead times beyond 48 hours. Although AFDs primarily align with GFS forecasts, they are sometimes based on other models as well, which diverge as the lead time increases due to the chaotic nature of the atmosphere. The spatial domain is cropped from 15$^{\circ}$N to 65$^{\circ}$N and 60$^{\circ}$W to 160$^{\circ}$W. We plot mean 2-meter temperature anomalies in shaded contours along with mean 500mb geopotential height in line contours and 850mb wind velocity in the form of wind barbs. A yellow 5$^{\circ}$ $\times$ 5$^{\circ}$ box is centered on the location of each discussion that is paired to an image. 

Keeping the goal of matching text that can be represented in the images, we preprocessed the information in the AFDs for the experiments. AFDs are often informed by GFS forecasts, but can use information from a variety of different models. We remove sentences including any reference to other forecast models or synoptic pressure systems that do not exist at the scale or vertical level that the images show. We also only use sentences including keywords that do describe synoptic pressure systems and temperature anomalies including: \emph{trough, ridge, low pressure, high pressure, cold front, and warm front}. The full list of inclusion and exclusion keywords is included in the appendix. When evaluating synoptic phenomena with lead times up to 48 hours, we can also expect robust consistency of large-scale patterns. The information described in AFDs occurs chronologically, often including the day of the week on which different phenomena occur. Once a day of the week that is more than two days after the issue day is mentioned, all text that follows is removed to stay consistent with the forecast lead times. 

\subsection{Baselines}
\label{sec:Baselines}

We used four different baselines to add perspective to the results. Gemini-3.1 Pro  is used as a baseline for current state-of-the-art models. The nearest neighbor baseline created by first computing the SSIM scores between each test set image and all training set images. The text paired with the training set image with the highest SSIM score is used as the text for each sample in the baseline. The climatology is created by choosing text from a random sample in the training set from the same month and location as each sample in the test set. The blind LLM baseline strips the image tokens from test before running LLaMA-3.2-11B.

\subsection{Synoptic Phenomena Alignment and Coverage Evaluation (\metric) metric}
\label{sec:metric}

\begin{figure*}
    \centering
    \includegraphics[width=0.81\textwidth]{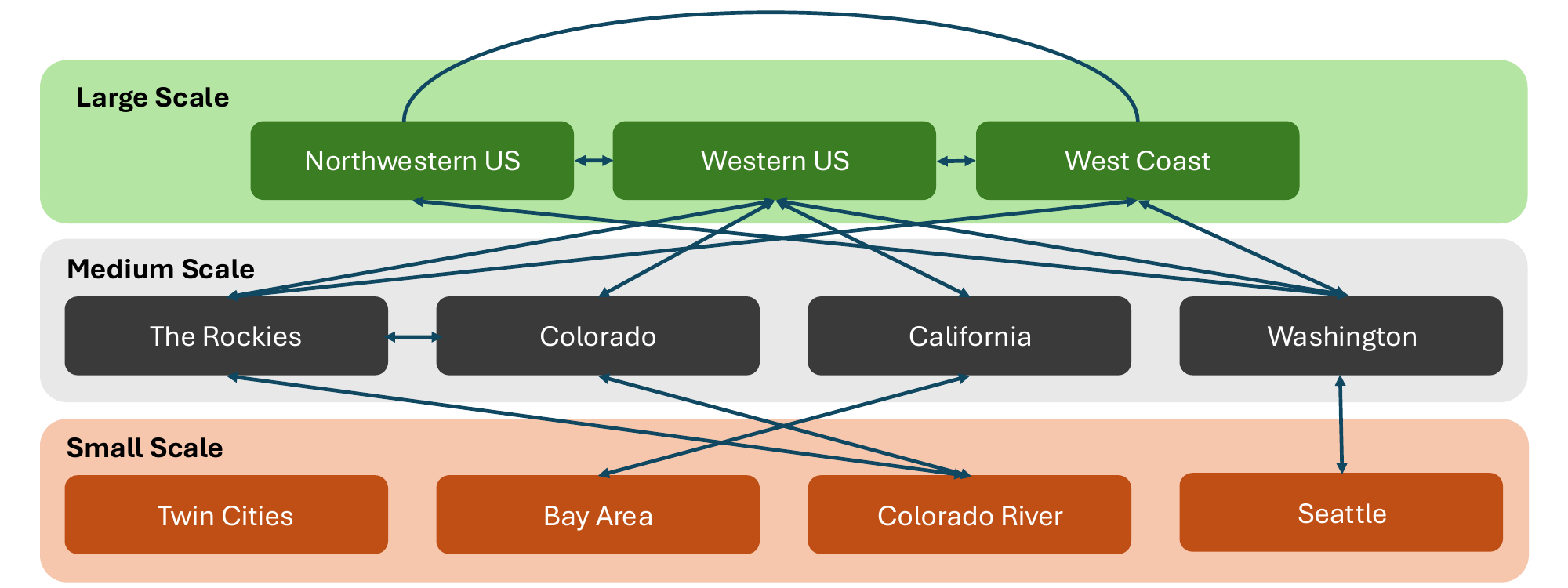}
    \caption{Several examples of matching large- (green), medium- (purple), and small-scale (orange) location keywords. Blue lines indicate the potential matches that these locations would make if found in the predicted or reference text}
    \label{fig:hierarchy}
\end{figure*}

Traditional metrics for text evaluation are not sufficient for text describing atmospheric processes because they do not include any physical reasoning. The purpose of \metric is to evaluate the ability of the VLM to generate text that predicts the correct phenomenon in the target location. Evaluating location accuracy from text can be a complex task because many terms may differ entirely, but represent similar locations (\eg The Rockies vs. Colorado) and the precision of location terms may be more or less acceptable depending on the scale of the phenomenon (\eg ``trough moving through the Denver area'' vs ``trough moving through Colorado'' is preferred more than ``rain in the Denver area'' vs. ``rain in Colorado''. To accommodate these relations, we introduce a spatial hierarchy tree specifically created for evaluating text describing synoptic phenomena. The hierarchy comprises three levels that capture terms commonly used to detail the locations of synoptic weather phenomena in North America (see Fig.~\ref{fig:hierarchy} for an example of a small portion of the hierarchy tree).

\metric can be applied towards various types of synoptic weather phenomena that have positive and negative phase categorizations (\eg high/low pressure systems, warm/cold fronts, wet/dry events). Next, to demonstrate the methodology of \metric, we use pressure systems as an example. We begin \metric by matching pressure terms to locations. If `low pressure', `high pressure', `trough' (excluding instances of `shortwave trough'), or `ridge' exist in the text, the term is first matched to any location that follows in the sentence. For a sample of text, this creates a list of pressure systems matched to different locations. Pressure terms are then matched to other instances for locations that are equal or related (see connections in Figure 2). Locations with a common relative can also match with the exception of those that cover a large portion of the continent (see Appendix). This logic creates groups of pressure objects for both the generated and the AFD text sample. Objects can be grouped for single discussions (\metric-local) or aggregated across all locations (\metric-aggregate) for a single forecast. If any object from a group has the same pressure sign and a location related to any object in the corresponding text sample, the object groups form a match. We consider `trough' to be synonymous with `low pressure', and `ridge' to be synonymous with `high pressure'. The match score, $s_m$, is a measure of the quality of the matches and is defined below:
\begin{equation}
s_{m} = 1 - \left|{\frac{m^{\text{pred}}_{L}}{m^{\text{pred}}_{L}+m^{\text{pred}}_{H}}} - \frac{m^{\text{ref}}_{L}}{m^{\text{ref}}_{L}+m^{\text{ref}}_{H}}\right|,
\end{equation}
where $m^{\text{pred}}_{L}$ is the number of matching \emph{negative phase} objects in the predicted text, $m^{\text{ref}}_{L}$ is the number of matching \emph{negative phase} objects in the reference text, $m^{\text{pred}}_{H}$ is the number of matching \emph{positive phase} objects in the predicted text, and $m^{\text{ref}}_{H}$ is the number of matching \emph{positive phase} objects in the reference text.  The coverage ratio, $r_{c}$, is the percentage of pressure objects that are included in matches and is defined as:
\begin{equation}
r_{c} = \frac{\sum((m^{\text{pred}}_{L} + m^{\text{pred}}_{H}) + (m^{\text{ref}}_{L} + m^{\text{ref}}_{H}))}{n_{L}+n_{H}},
\end{equation}
where $n_L$ is the total number of \emph{negative phase} objects and $n_H$ is the total number of \emph{positive phase} objects in the combined predicted and observed text. The coverage ratio is designed to penalize the generated text for creating hallucinations and for failing to mention key terms. The final \metric score is calculated as the product of the matching score and coverage ratio, \ie $s = s_m \cdot r_c$.

\section{Results}

\begin{table}
\centering
\resizebox{\textwidth}{!}{
\begin{tabular}{lccccc}
\toprule
  Model  & Bertscore &  ROUGE-L & METEOR & F1 & Gemini-2.5-Flash\\
\midrule
   LLaVA-v1.5-7B-base  & .7143\std{.0001} & .1091\std{.0001} & .1327\std{.0001} & .1471\std{.0002} & .1565\std{.0031}\\
   LLaVA-v1.5-13B-base  & .7192\std{.0001} & .1086\std{.0001} & .1376\std{.0001} & .1498\std{.0002} & .1896\std{.0032}\\
    LLaMA-3.2-11B-Vision-base  & .7051\std{.0001} & .1043\std{.0002} & .1212\std{.0002} & .1313\std{.0002} & .2623\std{.0041}\\
    Qwen2.5-VL-7B-base  & .7136\std{.0001} & .1083\std{.0001} & .1447\std{.0001} & .1591\std{.0003} & .2566\std{.0035}\\
    \midrule
   LLaVA-v1.5-7B-LoRA  & .7814\std{.0002} & .1629\std{.0003} & .1353\std{.0003} & .1943\std{.0004} & .2426\std{.0048}\\
   LLaVA-v1.5-13B-LoRA  & .7858\std{.0002} & .1663\std{.0003} & .1407\std{.0003} & .2010\std{.0004} & .2472\std{.0052}\\
   LLaMA-3.2-11B-Vision-LoRA  & .7773\std{.0001} & .1530\std{.0003} & .1174\std{.0003} & .1867\std{.0003} & .2757\std{.0047}\\
   Qwen2.5-VL-7B-LoRA  & \textbf{.7927}\std{.0002} & \textbf{.1681}\std{.0003} & .1601\std{.0004} & \textbf{.2184}\std{.0004} & .2824\std{.0053}\\
   \midrule
   Gemini-3.1-Pro  & .6913\std{.0004} & .0703\std{.0005} & .1456\std{.0008} & .1030\std{.0009} & \textbf{.4185\std{.0071}}\\
   Nearest Neighbor  & .7662\std{.0001} & .1194\std{.0002} & .1395\std{.0002} & .1748\std{.0003} & .1785\std{.0047}\\
   Climatology  & .7776\std{.0001} & .1434\std{.0002} & \textbf{.1632}\std{.0003} & .2035\std{.0003} & .2358\std{.0049}\\
   LLama-3.2-11B (blind)  & .7257\std{.0001} & .1101\std{.0002} & .1503\std{.0002} & .1544\std{.0003} & .2574\std{.0054}\\
 \bottomrule
 \end{tabular}
 }
 \label{tab:table1}\vspace{0.02in}
\caption{Traditional metrics for base models, finetuned models, and baselines are shown. We calculate the mean and standard error of the mean Bertscore, ROUGE-L, METEOR, F1, and LLM-judge scores across samples within the test set. The range of possible scores for all metrics is 0 to 1, with 1 being a perfect score}
 \end{table}

\xhdr{Traditional Metrics are not Sufficient} Base model versions for LLaVA-v1.5-7B, LLaVA-v1.5-13B, Qwen2-VL-7B, and LLaMA-3.2-11B-Vision were run on the test set. We first evaluate each of the four base models, finetuned models, and baselines on traditional skill metrics (Table 2). LoRA finetuning improved performance on all traditional metrics except METEOR for LLaMA-3.2-11B-Vision. The climatology baseline had the highest METEOR score, while finetuned Qwen2.5-VL-7B had the highest Bertscore, ROUGE-L, and F1 scores. Gemini-3.1-Pro had the lowest Bertscore, ROUGE-L score, METEOR score, and F1 score while having the highest LLM-as-a-judge score. The climatology baseline used text from real NWS AFDs issued from the same exact station, which may have allowed it to have relatively high scores for traditional metrics by using similar discussion styles. Despite slight variations in the performance of all models, all scores were low, indicating that the language in the generated discussions had little overlap with the NWS discussions. The traditional scores cannot determine the models' ability to discuss the primary relevant features in the images, which creates a need for further evaluation.

\begin{table}
\centering
\resizebox{\textwidth}{!}{
\begin{tabular}{lccccccc}
\toprule
  Model  & $s_{s}^{\text{loc}}$ &  $s_{m}^{\text{loc}}$ & $r_{c}^{\text{loc}}$ & $s_{s}^{\text{agg}}$ &  $s_{m}^{\text{agg}}$ & $r_{c}^{\text{agg}}$ \\
\midrule
   LLaVA-v1.5-7B-base  & .0976\std{.0008} & .4592\std{.0056} & .2158\std{.0013} & .4177\std{.0008} & .5039\std{.0056} & .9467\std{.0013} \\
   LLaVA-v1.5-13B-base  & .1553\std{.0007} & .4404\std{.0061} & .3615\std{.0012} & .4698\std{.0007} & .5044\std{.0061} & .9317\std{.0012} \\
    LLaMA-3.2-11B-Vision-base  & .1194\std{.0008} & .6296\std{.00013} & .1936\std{.0012} & .6492\std{.0008} & .6690\std{.0013} & .9648\std{.0012} \\
    Qwen2.5-VL-7B-base  & .0985\std{.0018} & .5072\std{.0032} & .1950\std{.0031} & .5831\std{.0018} & .6221\std{.0032} & .9279\std{.0031} \\
    \midrule
   LLaVA-v1.5-7B-LoRA  & .2235\std{.0015} & .6575\std{.0023} & .3364\std{.0018} & .6486\std{.0015} & .6714\std{.0023} & .9635\std{.0018} \\
   LLaVA-v1.5-13B-LoRA  & .2604\std{.0015} & .6601\std{.0021} & \textbf{.3900}\std{.0018} & .6410\std{.0015} & .6629\std{.0021} & .9648\std{.0018} \\
   LLaMA-3.2-11B-Vision-LoRA  & .2424\std{.0016} & .6554\std{.0023} & .3654\std{.0019} & .6091\std{.0016} & .6397\std{.0023} & .9488\std{.0019} \\
   Qwen2.5-VL-7B-LoRA  & \textbf{.2626}\std{.0015} & \textbf{.6744}\std{.0020} & .3845\std{.0018} & .6294\std{.0015} & .6498\std{.0020} & .9663\std{.0018} \\
   \midrule
   Gemini-3.1-Pro  & .2005\std{.0037} & .5825\std{.0034} & .3445\std{.0058} & .6685\std{.0037} & .6775\std{.0034} & \textbf{.9864}\std{.0058} \\
   Nearest Neighbor  & .1569\std{.0010} & .5485\std{.0015} & .2866\std{.0015} & .6187\std{.0010} & .6393\std{.0015} & .9661\std{.0015} \\
   Climatology  & .2009\std{.0016} & .5655\std{.0012} & .3554\std{.0018} & \textbf{.6808}\std{.0016} & \textbf{.6900}\std{.0012} & .9856\std{.0018} \\
   Llama-3.2 (blind)  & .1745\std{.0010} & .4967\std{.0014} & .3542\std{.0017} & .5690\std{.0010} & .5794\std{.0014} & .9813\std{.0017} \\

 \bottomrule
 \end{tabular}
 }
 \label{tab:table1}\vspace{0.02in}
\caption{SPACE was used for synoptic pressure system evaluation for base models, finetuned models, and baselines. We calculate the mean and standard error of the SPACE-local and SPACE-aggregate scores ($s_{s}$), match scores ($s_{m}$), and coverage ratios ($r_{c}$) within the test set. The range of possible scores for all metrics is 0 to 1, with 1 being a perfect score}
 \end{table}

\xhdr{\metric scores can Reduce Evaluation Uncertainty} Although \metric could be used for various types of weather phenomena, we demonstrate the utility of \metric on pressure systems in this study. Here, we show how \metric scores help can the user understand how each model performs in identifying the correct relevant pressure systems for the area and referring to them in the correct locations. \metric-aggregate uses all locations at a given time to create a large sample size of pressure objects matched to specific locations for evaluation. It is useful for understanding the model's ability to determine the general large-scale pressure features that occur in the forecast. \metric-local uses a single location at a given time to create a much smaller sample size of pressure objects but can be useful for understanding the model's ability to discuss the features impacting a specific location. All \metric scores for pressure systems are shown in Table 3. \metric-aggregate scores are considerably higher than \metric-local scores because the increased sample size of text increases the chances that relevant locations for synoptic phenomena are mentioned in both the generated text and the AFDs. Theoretically, a model that always randomly guesses between high pressure and low pressure would have a  match score of roughly 0.5 without any knowledge of climatology, which indicates that the finetuned models likely all have some skill in distinguishing between high and low pressure. The local coverage ratios ($r_{c}$) are lower than the match scores for every model, demonstrating that predicting the phenomena in the correct location is a more challenging task for these VLMs than accurately describing the correct polarity of synoptic phenomena. 

The Qwen2.5-VL-7B fine-tuned model had the highest \metric-local scores while Climatology had the highest \metric-aggregate scores. The fine-tuned models had higher \metric-local scores than all of the base models and baselines. The \metric-aggregate scores for the fine-tuned models were generally higher than those of the base models with the exception of LLaMA-3.2-11B-Vision. The improvement of \metric-aggregate scores over base models is impacted by both improved coverage ratios and match scores. The \metric scores often agree with the traditional metrics, with the fine-tuned Qwen-VL-7B configuration generally having the highest scores. One clear exception is Gemini-3.1-Pro, which has some of the lowest scores when evaluated on traditional metrics and fairly high \metric scores. The \metric scores and traditional metrics also suggest that increasing the parameters from 7 billion to 13 billion in LLaVA does not drastically improve generated discussion quality and the fine-tuning approach is far more effective. 

\xhdr{Case Studies Illustrate the Advantages of \metric} An example of the evaluation of both \metric and traditional metrics in two different cases is shown in Figure 3. In Case 1, there is a high pressure system over the southwestern United States, which is the most influential pressure system relevant to the station location (Tuscon, Arizona). The NWS discussion mentions high pressure in the region, and the generated VLM text also discusses high pressure systems in the same region. This leads to a perfect \metric-local score for pressure systems. The traditional metrics for this case were generally close to average for this model. In Case 2, there is a high pressure system over much of the southeastern US, which is the most influential pressure system to the station location (Columbia, South Carolina). The NWS discussion mentions a ridge in the forecast area while the VLM's generated text discusses a trough in the southeastern United States. This leads to a \metric score of 0. Despite having a \metric-local score of 0, the traditional metrics were unable to show a clear difference in the quality of both cases. Case 2 also shows some instances of hallucinations in which the model discusses phenomena that it could not possibly see based on the image (e.g. 1 inch of snow), which does not penalize the \metric score because \metric only evaluates one type of phenomenon at a time. This case study demonstrates a clear example of the reliability of \metric scores relative to traditional metrics for the evaluation of individual types of phenomena in weather discussions. Several additional cases are shown in the Appendix.

\begin{figure}[h!]
 \centerline{\includegraphics[width=0.9\textwidth]{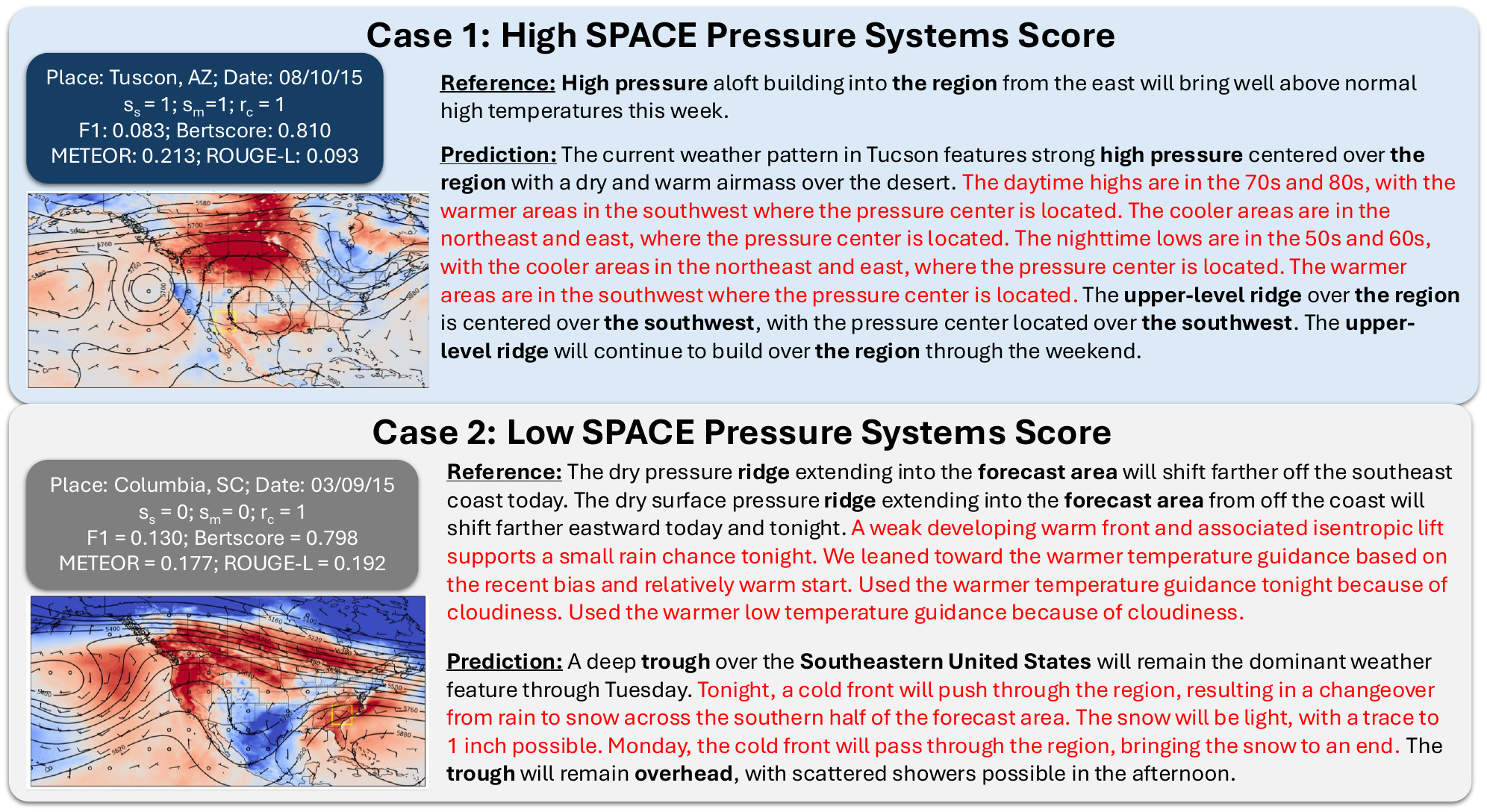}}
  \caption{Two cases demonstrating differences between \metric scores and traditional skill metrics. The reference text samples are filtered NWS AFDs and the prediction text samples are generated from the finetuned version of LLaVA-v1.5-7B. The terms in bold are used to compute \metric scores for pressure systems. Sentences that are irrelevant to the \metric scores are shown in red}\label{Fig4}
\end{figure}

\section{Conclusion}

In this work, we introduce a dataset with more than one million samples that we use to train a Vision Language Model for discussion generation from images of weather forecasts across North America. We also introduce a skill metric (\metric) to evaluate the ability of language models to identify important synoptic features in the correct locations. Fine-tuning VLM projector weights with LoRA adapters leads to greater overall performance than the base models on both \metric and traditional metrics. Climatology and finetuned Qwen2.5-VL-7B had the highest scores in most of the metrics in this study, with Qwen2.5-VL-7B excelling more in describing features relevant to a precise location. The failure of any model to score higher than climatology in the \metric-aggregate scores demonstrates a key limiting factor in the ability of the models in this study to accurately describe synoptic features throughout the entire system. 

The data set used in this study pairs highly complex forecast discussions from the National Weather Service with images of weather forecasts. This study performed experiments on synoptic pressure systems as an example use case and focused entirely on the United States, which creates an opportunity for future work. Human expert evaluation for text describing various types of atmospheric phenomena would also be valuable for strengthening confidence in any current or future metrics. The size and complexity of the data set allow the user to perform experiments on other types of phenomena, specific locations or regions, spatial scales, and temporal ranges. Although the finetuned models often outperformed  the base models on \metric scores, improved model architectures and loss functions,  agentic frameworks, and video-language models could add additional benefits. 

Weather forecasts are highly complex, and complete descriptions of them therefore often require long and detailed discussions. None of the models used in this study come close to rivaling the reliability nor complexity of NWS AFDs, so the evaluation of text generation for weather and climate applications must begin with basic isolated tasks. \metric scores excel in capturing similarities in some of the most important details of text samples. They can help evaluate the ability of the models to demonstrate physical understanding of atmospheric processes rather than only learning common textual patterns. By determining the ability of the model to do so, the user can better understand when to stop training, which can help prevent overfitting. 

Although \metric scores are a useful method for evaluating text for multimodal language models for weather and climate applications, they do not fully capture the level of quality in generated forecast discussions, and additional metrics that evaluate text would help create a more comprehensive evaluation framework. Future improvements could involve temporal accuracy (more relevant for longer lead times), system evolution (more relevant for models that can handle videos), and vertical levels (more relevant for models that can handle more channels).

\paragraph{Acknowledgments}

The authors thank the reviewers of this work for their dedication towards supporting the research community and providing comments that were critical towards improving the manuscript.

\paragraph{Funding Statement}
This work was funded by the University of Virginia Environmental Institute Climate Fellows Program Grant DN002057. 

\paragraph{Competing Interests}
None

\paragraph{Data Availability Statement}
The code used to conduct this study can be found here: \url{github.com/timbhiggins/Synoptic-Bench}. The dataset can be found here: \url{https://huggingface.co/datasets/Aikyam-Lab/Synoptic-Bench}

\paragraph{Ethical Standards}
The research meets all ethical guidelines, including adherence to the legal requirements of the study country.

\paragraph{Author Contributions}
T.H., A.M., and C.A. contributed new analytic metrics and wrote the manuscript. T.H. wrote all the codes, retrieved, processed, and harmonized datasets, and performed the analyses for technical validation of the new resource. A.M. and C.A. conceived the study.

\bibliography{references}

\appendix
\renewcommand{\thefigure}{A\arabic{figure}}
\setcounter{figure}{0}
\section*{Appendix}

\section{Experimental details}

\xhdr{Implementation Details} We fine-tune each base model by training for roughly 1-3 epochs on one NVIDIA H200 GPU before running it on the test set. We use 2016-2022 for the training set, 2023 as the validation set, and 2024-2025 as the test set. The projector weights were fully fine-tuned, while Low-Rank Adaptation (LoRA) was used to train a small portion of the attention mechanism in the language model (<2\%). We apply weight decay and dropout to the LoRA adapter branches to prevent overfitting. In addition, early stopping is used by computing \metric-local scores for the validation set every 5000 steps. 
\vspace{0.05in}

\xhdr{Prompt Details} Base model and finetuned model versions for LLaVA-v1.5-7B, LLaVA-v1.5-13B, Qwen2-VL-7B, and LLaMA-3.2-11B-Vision were run on the test set. Each model uses the following prompt to generate discussions (see Fig. 1): "Analyze these weather charts showing maximum precipitation over the 3-day period (shaded contours), 500 mb geopotential height (unshaded contours) and 850 mb wind velocity (wind barbs) for the <[City, State]> forecast region and generate a detailed forecast discussion of the large-scale features relevant to the area. The forecast region is shown in the yellow box".

\xhdr{Additional Preprocessing Details} The images used in this study use the temporal mean of lead times ranging from 3 hours to 48 hours in 3 hour increments. The anomalies are calculated from monthly means ranging across the entire dataset. The month chose for climatology is always the same month that the AFD is issued. To create forecast images, 2-m temperature anomaly is capped between -5° and 5°C. 500 mb geopotential height contours are drawn at intervals of 60m and 850mb wind velocity barbs are plotted every 20th grid point to avoid visual clutter. 

To filter AFDs, we only extract sentences containing the following keywords: \emph{trough, trof, the low, this low, upper level low, low pressure, low-pressure, upper low, cyclone, closed low, cut-off low, troughing, ridge, the high, this high, upper level high, high pressure, high-pressure, upper high, anticyclone, blocking, ridging, cold front, warm front, warmer, cooler, and freezing}. We also exclude sentences with the following model keywords: \emph{ECMWF, EURO, HRRR, ECCC, CMC, GEM, NAM, UKMET, ICON, RAP, SREF, HREF,} and synoptic pressure system keywords: \emph{shortwave, short-wave, sfc trough, surface trough, sfc ridge, surface ridge, surface low, sfc low, surface high, sfc high} to increase the relevancy of NWS text in the use case presented in this study. 

\xhdr{Parameter Settings} The same settings for the finetuning parameters were used for LLaVA-v1.5-7B, LLaVA-v1.5-13B, Qwen2-VL-7B, and LLaMA-3.2-11B-Vision. The learning rate was set at 2e-5, the batch size was set to 16, the LoRA rank was set to 8, and the LoRA alpha was set to 16. LoRA was applied to all linear modules within the attention and multi-layer perceptron blocks. AdamW was used as an optimizer. Image preprocessing was handled natively by each model's respective Hugging Face AutoProcessor. The images were converted to RGB and dynamically resized, padded, and normalized according to the base model's default vision encoder requirements (e.g., SigLIP for LLaVA/Llama and dynamic resolution for Qwen2.5-VL) before being tokenized.

\xhdr{Data Quality Control} Each text sample is matched to the closest GFS run. The initial raw dataset contained 1,367,041 samples. Text samples that are not within 12 hours of any GFS run are discarded. Pre-filtered samples in the raw dataset containing less than 30 words or greater than 200 words were also discarded. In addition, we removed samples containing text that suggested longer lead times beyond the 48-hour lead time post filtering (e.g. “Day 3”, “extended period”, “long term”). Quality control tests revealed zero samples containing text duplicates, unusual characters, absent image tokens, missing image files, or corrupted image files. Once all of the filtering specific to the example use case of this experiment was completed, the training set was reduced to 457,230 high-quality samples.

\xhdr{Extraction Details} We developed an automated Python pipeline to interface with the IEM JSON API, systematically querying and downloading historical AFDs from 117 NWS Weather Forecast Offices. The raw text files were extracted for the period spanning January 2019 to November 2025, yielding a comprehensive collection of chronologically ordered NWS discussions. To ensure strict temporal accuracy for multimodal pairing, we implemented a custom regular expression parser to extract the exact issuance timestamp directly from the raw text body of each product, converting all times to Coordinated Universal Time (UTC).

\xhdr{LLM as a Judge} Gemini-2.5-Flash was used as a judge for evaluation. The following text was used as the prompt for the judge: "Compare the generated forecast discussion to the ground truth forecast discussion. Evaluate how accurately the generated discussion captured the synoptic-scale features. Ignore minor differences in phrasing and focus on meteorological accuracy and completeness. Score the prediction from 0 to 1:

0: Completely incorrect or severely hallucinated.

.25: Major meteorological errors or missing critical synoptic features.

.5: Partially correct, but missing some context or containing minor errors.

.75: Mostly accurate and aligns well with the ground truth, minor omissions.

1: Excellent, highly accurate, and meteorologically sound compared to the ground truth.

\xhdr{Location Stop Nodes} The following locations are excluded from functioning as common relatives or matching to other locations via a common relative: "Canada", "CONUS", "Eastern Canada", "Central Canada", "Western Canada", "Eastern CONUS", "Western CONUS", "Central CONUS", "Central Plains", "Ohio Valley", "Great Lakes", "Central U.S.", "Eastern U.S.", "Western U.S.", "Central United States", "Eastern United States", "Western United States", "Central U", "Eastern U", "Western U", "Eastern US", "Western US", "Central US", "Midwest", "The Plains".

\xhdr{\metric Temperature Scores} SPACE scores for temperature are shown in Table 4. The following key terms for temperature polarity were used: "warm front", "warmer than average", "warmer than normal", "above average temperature", "cold front", "cooler than average", "cooler than normal", "colder than normal", and "below average temperature". The number of temperature terms found in both AFDs and generated text was dominated by "cold front" and "warm front" (>80\%). Fine-tuned Qwen2.5-VL-7B-LoRA had the highest \metric-local scores while fine-tuned LLaVA-v1.5-7B had the highest \metric-aggregate scores. The 48-hour temporal average that was used to create the images could lead to too much smoothing for the temporal scale at which fronts often take effect. Although \metric scores could be effective for synoptic temperature systems, altering training images and experimental setup to target temperature-related phenomena would improve the reliability of evaluation in future work.

\xhdr{Additional Example Cases} Three additional example cases are shown in Figures A1-A3. In Case 3, there are two matching instances of pressure systems and two unmatching instances of pressure systems, leading to a coverage ratio of 0.5. Of the matching instances, both indicated high pressure, resulting in a match score of 1 and a SPACE-local score of 0.5. In Case 4, the prediction described synoptic pressure systems, while the reference text did not (Short wave troughs are not considered synoptic pressure systems). This results in a SPACE-local score that defaults to 0. Although there was some accurate information in the generated text, it was not relevant enough to the area to be included in the AFD. The pressure systems described in the generated text could still impact SPACE-aggregate scores. In Case 5, all pressure systems were described in the same general area, leading to a coverage ratio of 1. The match score was imperfect due to a ridge that would impact the region soon after the trough. 

\xhdr{Training Set Size} The impact of the size of the training set on the  \metric scores is shown in Figure A4. The majority of the improvement in \metric scores occurs within the first half epoch of training.

\begin{table}
\centering
\resizebox{\textwidth}{!}{
\begin{tabular}{lccccccc}
\toprule
  Model  & $s_{s}^{\text{loc}}$ &  $s_{m}^{\text{loc}}$ & $r_{c}^{\text{loc}}$ & $s_{s}^{\text{agg}}$ &  $s_{m}^{\text{agg}}$ & $r_{c}^{\text{agg}}$ \\
\midrule
   LLaVA-v1.5-7B-base  & .0173\std{.0028} & .8278\std{.0526} & .0207\std{.0031} & .3447\std{.0028} & .5602\std{.0526} & .6017\std{.0031} \\
   LLaVA-v1.5-13B-base  & .0793\std{.0048} & \textbf{.9068}\std{.0125} & .0862\std{.0051} & .4229\std{.0008} & .6053\std{.0056} & .6843\std{.0013} \\
    LLaMA-3.2-11B-Vision-base  & .0671\std{.0025} & .7294\std{.0075} & .0886\std{.0030} & .3300\std{.0025} & .5683\std{.0075} & .5709\std{.0030} \\
    Qwen2.5-VL-7B-base  & .0491\std{.0032} & .8330\std{.0155} & .0577\std{.0035} & .3302\std{.0032} & .5478\std{.0155} & .5878\std{.0035} \\
    \midrule
   LLaVA-v1.5-7B-LoRA  & .1863\std{.0056} & .8839\std{.0075} & .2077\std{.0060} & \textbf{.4679}\std{.0056} & .5965\std{.0075} & .7718\std{.0060} \\
   LLaVA-v1.5-13B-LoRA  & .1862\std{.0063} & .8697\std{.0085} & .2098\std{.0067} & .4209\std{.0063} & .5870\std{.0085} & .7043\std{.0067} \\
   LLaMA-3.2-11B-Vision-LoRA  & .1726\std{.0055} & .8992\std{.0073} & .1892\std{.0057} & .4597\std{.0055} & \textbf{.6124}\std{.0073} & .7395\std{.0057} \\
   Qwen2.5-VL-7B-LoRA  & \textbf{.1886}\std{.0074} & .8660\std{.0058} & \textbf{.2128}\std{.0062} & .3992\std{.0074} & .5978\std{.0058} & .6529\std{.0062} \\
   \midrule
   Gemini-3.1-Pro  & .0397\std{.0050} & .7725\std{.0340} & .0408\std{.0057} & .3963\std{.0050} & .5135\std{.0340} & \textbf{.7732}\std{.0057} \\
   Nearest Neighbor  & .0585\std{.0036} & .7532\std{.0166} & .0753\std{.0041} & .3216\std{.0036} & .5607\std{.0166} & .5614\std{.0041} \\
   Climatology  & .0850\std{.0042} & .7728\std{.0148} & .1073\std{.0048} & .4042\std{.0042} & .5795\std{.0148} & .6894\std{.0048} \\
   Llama-3.2 (blind)  & .0912\std{.0048} & .8664\std{.0141} & .1043\std{.0051} & .4211\std{.0048} & .5885\std{.0141} & .7050\std{.0051} \\

 \bottomrule
 \end{tabular}
 }
 \label{tab:table1}\vspace{0.02in}
\caption{SPACE was used for synoptic temperature system evaluation for base models, finetuned models, and baselines. We calculate the mean and standard error of the SPACE-local and SPACE-aggregate scores ($s_{s}$), match scores ($s_{m}$), and coverage ratios ($r_{c}$) within the test set. The range of possible scores for all metrics is 0 to 1, with 1 being a perfect score}
 \end{table}

\begin{figure}[h!]
 \centerline{\includegraphics[width=0.9\textwidth]{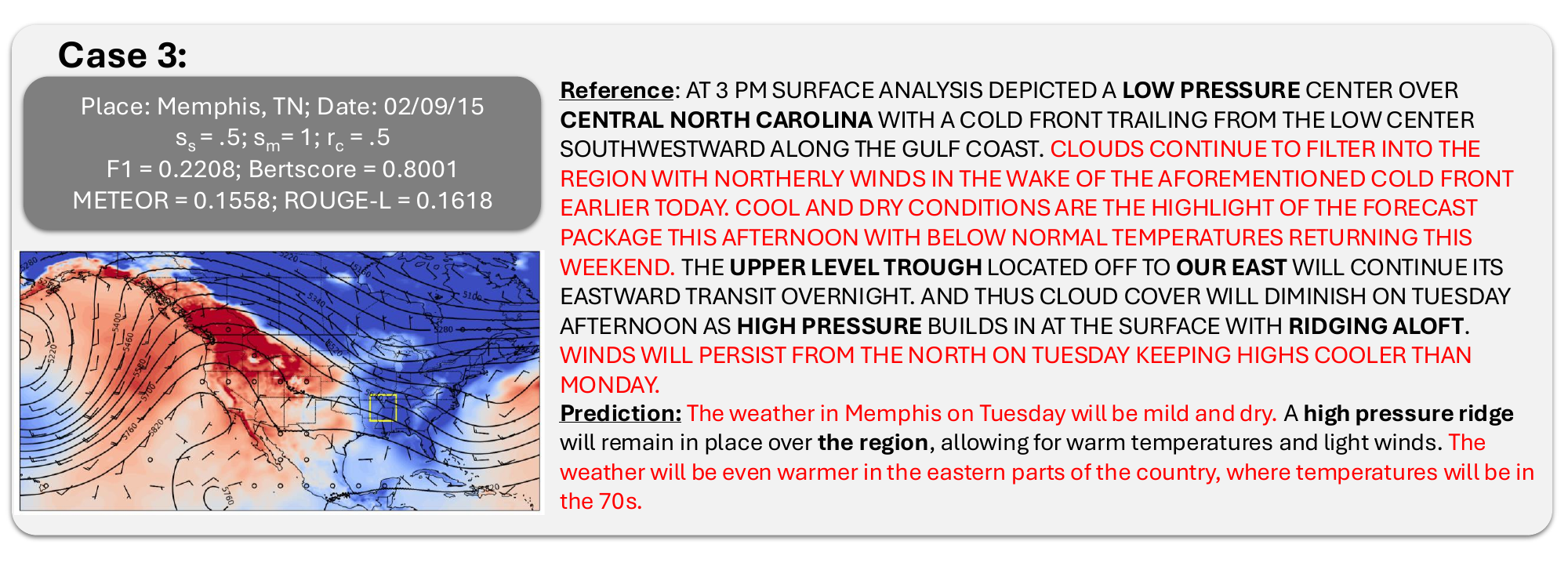}}
  \caption{Example Case 3}\label{Fig A1}
\end{figure}

\begin{figure}[h!]
 \centerline{\includegraphics[width=0.9\textwidth]{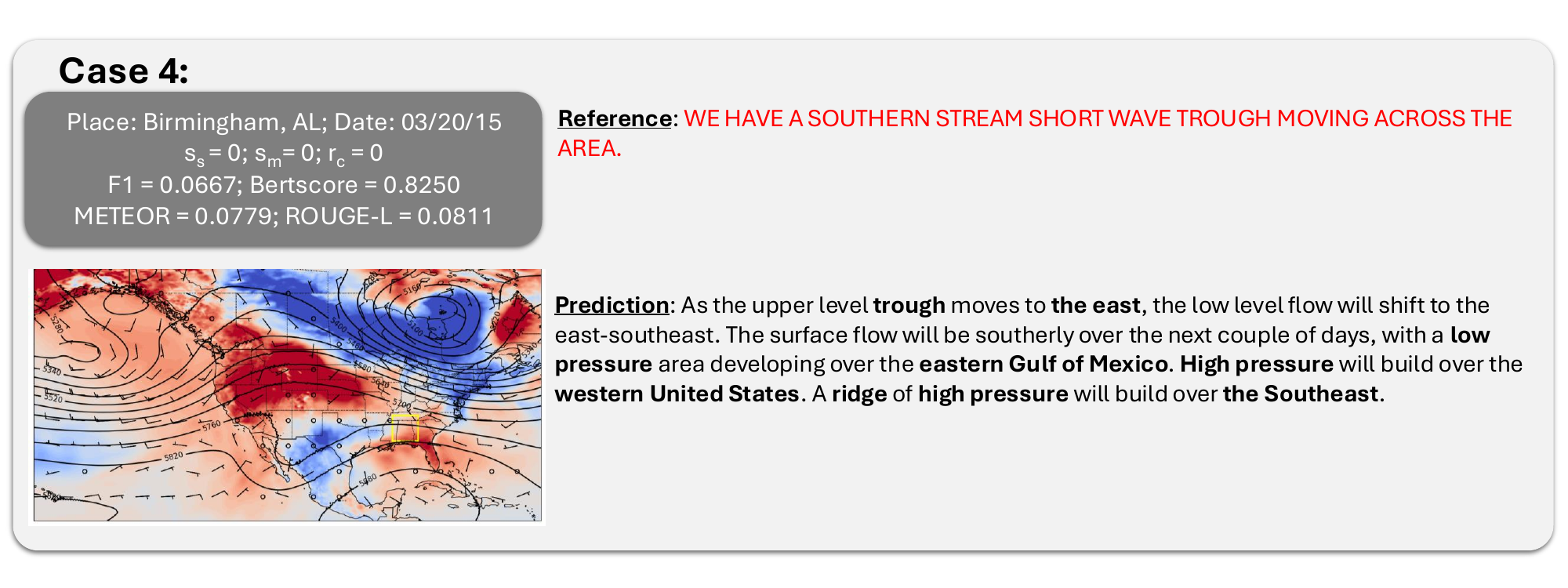}}
  \caption{Example Case 4}\label{Fig A2}
\end{figure}

\begin{figure}[h!]
 \centerline{\includegraphics[width=0.9\textwidth]{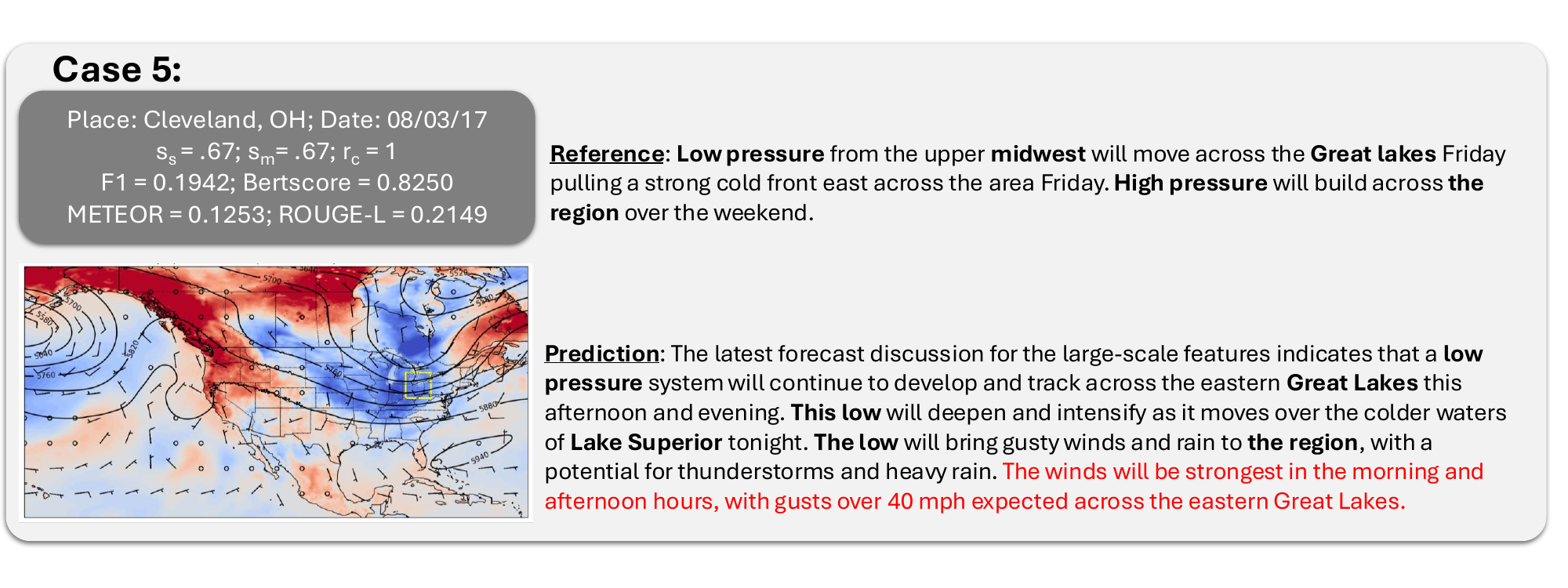}}
  \caption{Example Case 5}\label{Fig A3}
\end{figure}

\begin{figure}[h!]
 \centerline{\includegraphics[width=0.9\textwidth]{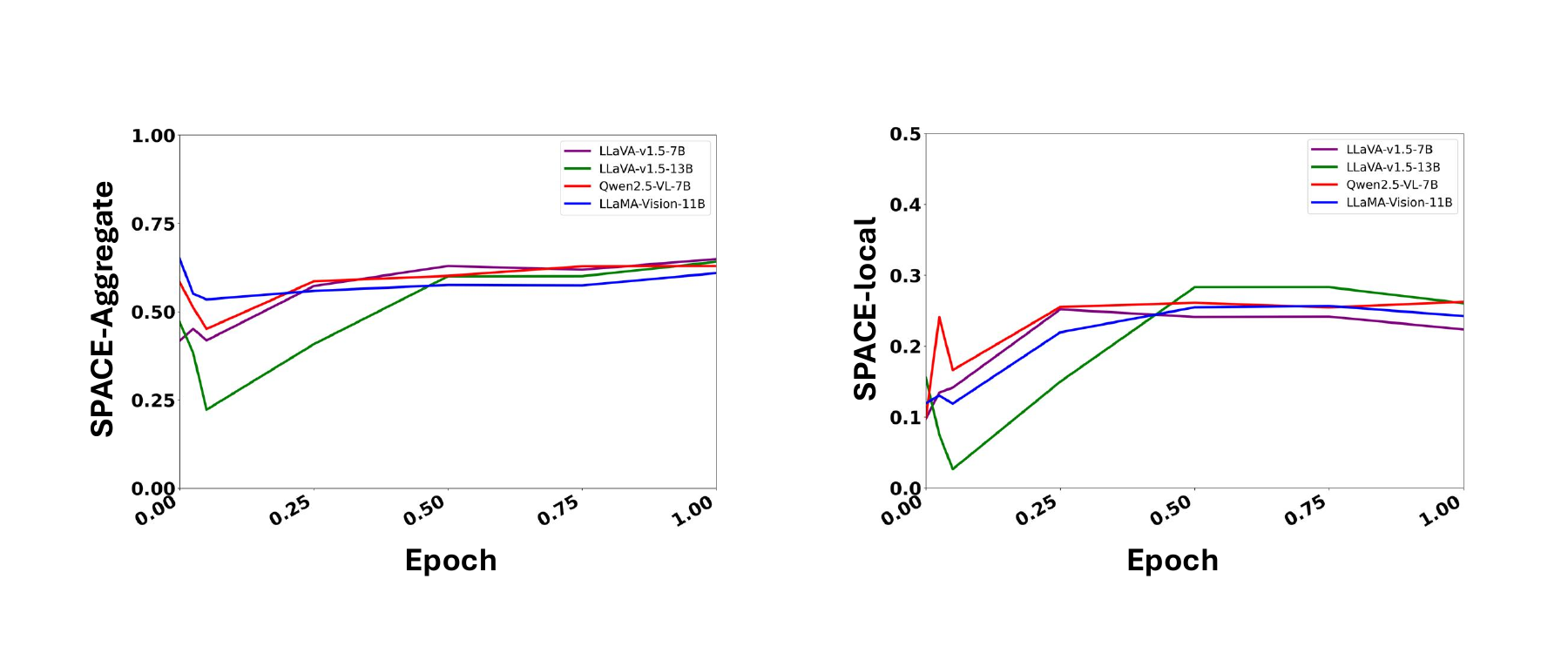}}
  \caption{Training Set Size}\label{Fig A4}
\end{figure}

\end{document}